\documentclass[a4,10pt]{article}

\usepackage{latexsym}

\usepackage{graphicx}
\usepackage{pstricks}
\usepackage{amssymb}
\usepackage{amsmath}
\usepackage{fullname}

\input{ntvit.def}

\begin{document}


\title{ \vspace{-2ex}
  Viterbi Algorithm Generalized for $n$-Tape Best-Path Search
}

\author{
  Andr\'e Kempe \vspace{2ex} \\
  Xerox Research Centre Europe ~--~ Grenoble Laboratory \\
  6 chemin de Maupertuis ~--~ 38240 Meylan ~--~ France \\
}

\date{
  March 9, 2006
}

\maketitle

\begin{abstract}
  We present a generalization of the Viterbi algorithm
  for identifying the path with minimal (resp. maximal) weight
  in a {\it $n$-tape weighted finite-state machine}\/ ($n$-WFSM),
  that accepts a given $n$-tuple of input strings $\aTuple{s_1,\ldots s_n}$.
  It also allows us to compile the best transduction of a given input $n$-tuple
  by a weighted $(n\!+\!m)$-WFSM (transducer) with $n$ input and $m$ output tapes.
  Our algorithm has a worst-case time complexity of
  $\complexity\left(\, |s|^n|E|\log|s|^n|Q| \,\right)$,
  where $n$ and $|s|$ are the number and average length of the strings in the $n$-tuple,
  and $|Q|$ and $|E|$ the number of states and transitions in the $n$-WFSM,
  respectively.
  A straight forward alternative,
  consisting in intersection followed by classical shortest-distance search,
  operates in $\complexity\left(\, |s|^n(|E|+|Q|)\log|s|^n|Q| \,\right)$ time.
\end{abstract}


\section{Introduction
  \label{sec:intro}}

The topic of this paper is situated in the areas of
{\it multi-tape}\/ or {\it $n$-tape weighted finite-state machines}\/ ($n$-WFSMs)
and shortest-path problems.


$n$-WFSMs
\cite{rabin+scott:1959,elgot+mezei:1965,kay:1987,harju+karhumaki:1991,kaplan+kay:1994}
are a natural generalization of the familiar
finite-state acceptors (one tape) and transducers (two tapes).
The $n$-ary relation defined by an $n$-WFSM is a weighted {\em rational\/} relation.
Finite relations are of particular interest since they
can be viewed as relational databases.
A finite-state transducer ($n=2$) can be seen as a database of string pairs,
such as $\aTuple{\WordD{spelling}, \WordD{pronunciation}}$ or
$\aTuple{\WordD{French word}, \WordD{English word}}$.
Unlike a classical database, a transducer may even define infinitely many pairs.
For example, it may characterize the pattern of the spelling-pronunciation
relationship in such a way that it can map even the spelling of an unknown word
to zero or more possible pronunciations (with various weights),
and vice-versa.
$n$-WFSMs have been used in the morphological analysis of Semitic languages,
to synchronize the vowels, consonants, and
templatic pattern into a surface form \cite{kay:1987,kiraz:2000}.


Classical shortest-path algorithms can be separated into two groups,
addressing either single-source shortest-path (SSSP) problems,
such as Dijkstra's algorithm \cite{dijsktra:1959}
or Bellman-Ford's \cite{bellman:1958,ford+fulkerson:1956},
or all-pairs shortest-path (APSP) problems,
such as Floyd-Warshall's \cite{floyd:1962,warshall:1962}.
SSSP algorithms determine a minimum-weight path from a source vertex
of a real- or integer-weighted graph to all its other vertices.
APSP algorithms find shortest paths between all pairs of vertices.
For details of shortest-path problems in graphs see \cite{pettie:2003},
and in semiring-weighted finite-state automata see \cite{mohri:2002b}.


\smallskip

We address the following problem:
in a given $n$-WFSM we want to identify the path with minimal (resp. maximal) weight
that accepts a given $n$-tuple of input strings $\aTuple{s_1,\ldots s_n}$.
This is of particular interest because it allows us also
to compile the best transduction of a given input $n$-tuple
by a weighted $(n\!+\!m)$-WFSM (transducer) with $n$ input and $m$ output tapes.
For this, we identify the best path accepting the input $n$-tuple on its input tapes,
and take the label of the path's output tapes as best output $m$-tuple.

A known straight forward method for solving our problem is
to intersect the $n$-WFSM with another one that contains a single path
labeled with the input $n$-tuple,
and then to apply a classical SSSP algorithm, ignoring the labels.
We show that such an intersection together with Dijkstra's algorithm have
a worst-case time complexity of
$ \complexity\left(\, |s|^n(|E|+|Q|)\log|s|^n|Q| \,\right) $,
where $n$ and $|s|$ are the number and average length of the strings in the $n$-tuple,
and $|Q|$ and $|E|$ the number of states and transitions of the $n$-WFSM,
respectively.

We propose an alternative approach with lower complexity.
It is based on the Viterbi algorithm
which is generally used for detecting the most likely path
in a {\it Hidden Markov Model}\/ (HMM)
for an observed sequence of symbols emitted by the HMM
\cite{viterbi:1967,rabiner:1990,manning+schuetze:1999}.
Our algorithm is a generalization of Viterbi's algorithm
such that it deals with an $n$-tuple of input strings rather than with a single input string.
In the worst case,
it operates in $\complexity\left(\, |s|^n|E|\log|s|^n|Q| \,\right)$ time.


\smallskip

This paper is structured as follows.
Basic definitions of weighted $n$-ary relations, $n$-WFSMs, HMMs, and the Viterbi algorithm
are recalled in Section~\ref{sec:prelim}.
Section~\ref{sec:vit-1tape}
adapts the Viterbi algorithm
to the search of the best path in a $1$-WFSM that accepts a given input string,
and Section~\ref{sec:vit-ntape} generalizes it
to the search of the best path in an $n$-WFSM that accepts an $n$-tuple of strings.
Section~\ref{sec:align}
illustrates our algorithm on a practical example,
the alignment of word pairs (i.e., $n\!=\!2$),
and provides test results that show a slightly higher than
$\complexity\left(\, |s|^2 \,\right)$ time complexity.
The above mentioned classical method for solving our problem
is discussed in Section~\ref{sec:alternatives}.
Section~\ref{sec:conclusion}
concludes the paper.


\section{Preliminaries
  \label{sec:prelim}}

We recall some definitions about
$n$-ary weighted relations and their machines,
following the usual definitions for multi-tape
automata \cite{elgot+mezei:1965,eilenberg:1974},
with semiring weights added just as for acceptors and transducers
\cite{kuich+salomaa:1986,mohri+al:1998}.
For more details see \cite{kempe+champarnaud+eisner:2004a}.
We also briefly recall Hidden Markov Models and the Viterbi algorithm,
and point the reader to
\cite{viterbi:1967,rabiner:1990,manning+schuetze:1999}
for further details.


\subsection{Weighted $n$-ary relations}

A weighted $n$-ary relation is a function from $(\Sigma^*)^n$ to
$\srSetK$, for a given finite alphabet $\Sigma$ and a given weight
semiring $\srK = \aTuple{\srSetK, \srPlus, \srTimes, \srZero,
  \srOne}$.  A relation assigns a weight to any
$n$-tuple of strings.  A weight of $\srZero$ can be interpreted as
meaning that the tuple is not in the relation.
We are especially interested in {\it rational} (or {\it regular})
$n$-ary relations, i.e. relations that can be encoded by $n$-tape
weighted finite-state machines, that we now define.

We adopt the convention that variable names referring to $n$-tuples of
strings include a superscript $\tapnum{n}$.  
Thus we write $s\tapnum{n}$ rather than
${\mathop{s}\limits^\rightarrow}$ 
for a tuple of strings $\aTuple{s_1, \dots  s_n}$.  
We also use this convention for the names of 
objects that contain $n$-tuples of strings,
such as $n$-tape machines and their transitions and paths.


\subsection{Multi-tape weighted finite-state machines}

An {\it $n$-tape weighted finite-state machine} (WFSM or $n$-WFSM)
$A\tapnum{n}$ is defined by a six-tuple
$A\tapnum{n} = \aTuple{\Sigma, Q, \srK, E\tapnum{n}, \wgtInit, \wgtFin}$,
with
$\Sigma$ being a finite alphabet,
$Q$ a finite set of states,
$\srK\!=\!\aTuple{\srSetK,\srPlus,\srTimes,\srZero,\srOne}$ the
semiring of weights, 
$E\tapnum{n}\!\subseteq ( Q\times (\Sigma^*)^n \times \srSetK \times Q )$
		a finite set of weighted $n$-tape transitions, 
$\wgtInit : Q \rightarrow \srSetK$ a function that assigns initial
weights to states, 
and $\wgtFin : Q \rightarrow \srSetK$ a function that assigns final
weights to states.

Any transition $e\tapnum{n}\!\in\! E\tapnum{n}$ has the form
$e\tapnum{n}\!=\!\aTuple{\eSrc,\lab\tapnum{n},w,\eTrg}$.
We refer to these four components as the transition's source state
$\eSrc(e\tapnum{n})\!\in\!Q$, its label
$\lab(e\tapnum{n})\!\in\!(\Sigma^*)^n$, its weight
$w(e\tapnum{n})\!\in\!\srSetK$, and its target state
$\eTrg(e\tapnum{n})\!\in\!Q$.  
We refer by $E(q)$ to the set of out-going transitions of a state $q\!\in\!Q$
~(with $E(q)\!\subseteq\!E\tapnum{n}$).

A {\it path}\/ $\path\tapnum{n}$ of length $k \geq 0$
is a sequence of transitions
$e_1\tapnum{n} e_2\tapnum{n} \cdots e_k\tapnum{n}$
such that $\eTrg(e_i\tapnum{n})\!=\!\eSrc(e_{i+1}\tapnum{n})$
for all $i\!\in\!\aRange{1, k\!-\!1}$.
The label of a path is the element-wise concatenation of 
the labels of its transitions.
The weight of a path $\path\tapnum{n}$ is
%
\begin{equation}
  w(\path\tapnum{n})	\DefAs
	\wgtInit(\eSrc(e_1\tapnum{n})) \srTimes
	\left(\srBigTimes_{j\in\aRange{1,k}}
		\spc{-1ex}w\left(e_j\tapnum{n}\right)\right) \srTimes
	\wgtFin(\eTrg(e_k\tapnum{n}))
\end{equation}


\noindent
The path is said to be {\it successful}, and to {\it accept}
its label, if $w(\path\tapnum{n})\neq\srZero$.


\subsection{Hidden Markov Models}

A {\it Hidden Markov Model}\/ (HMM) is defined by a five-tuple
$\aTuple{\Sigma,Q,\hmIniVec,\hmTraMtx,\hmOutMtx}$, where
$\Sigma\!=\!\aSet{\sigma_k}$ is the output alphabet,
$Q\!=\!\aSet{q_i}$ a finite set of states,
$\hmIniVec\!=\!\aSet{\hmIniPrb_i}$ a vector of initial state probabilities
  $\hmIniPrb_i = p(x_1\!=\!q_i) : Q\rightarrow\aRange{0,1}\,$,~
$\hmTraMtx\!=\!\aSet{\hmTraPrb_{ij}}$ a matrix of state transition probabilities
  $\hmTraPrb_{ij} = p(x_t\!=\!q_j | x_{t-1}\!=\!q_i) : Q\!\times\!Q\rightarrow\aRange{0,1}\,$,~
and $\hmOutMtx\!=\!\aSet{\hmOutPrb_{jk}}$ a matrix of state emission probabilities
  $\hmOutPrb_{jk} = p(\hmPthOut_t\!=\!\sigma_k | x_t\!=\!q_j) : Q\!\times\!\Sigma\rightarrow\aRange{0,1}\,$.
A {\it path}\/ of length $T$ in an HMM is a non-observable (i.e., hidden) state sequence
$\hmStaSeq = \hmPthSta_1\cdots\hmPthSta_T$,
emitting an observable output sequence
$\hmOutSeq = \hmPthOut_1\cdots\hmPthOut_T$
which is a probabilistic function of $\hmStaSeq$.
%


\subsection{Viterbi Algorithm}

\begin{figure}[htb]
  \begin{center}
    \includegraphics[scale=0.5,angle=0]{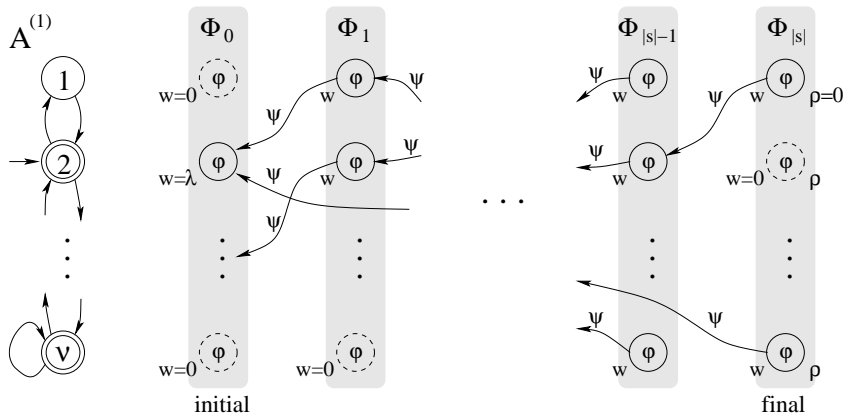}
    \caption{Modified trellis for $1$-tape best-path search
	\label{fig:vit-1tape}}
  \end{center}
\end{figure}

The {\it Viterbi algorithm}\/
finds the most likely path
$\widehat\hmStaSeq = \argmax_{\hmStaSeq} p(\hmStaSeq | \hmOutSeq, \mu)$
for an observed output sequence $\hmOutSeq$
and given model parameters $\mu=\aTuple{\hmIniVec,\hmTraMtx,\hmOutMtx}$,
using a trellis similar to that in Figure~\ref{fig:vit-1tape}.
It has a $\complexity(T\, |Q|^2)$ time and a $\complexity(T\, |Q|)$ space complexity.


\section{$1$-Tape Best-Path Search
  \label{sec:vit-1tape}}

The Viterbi algorithm
\cite{viterbi:1967,rabiner:1990,manning+schuetze:1999}
can be easily adapted for searching for the best of all paths of a $1$-WFSM, $A\tapnum{1}$,
that accept a given input string.
We use a notation that will facilitate the subsequent generalization 
of the algorithm to $n$-tape best-path search (Section~\ref{sec:vit-ntape}). 
Only the search for the path with minimal weight is explained.
An adaptation to maximal weight search is trivial.


\subsection{Structures}

We use a reading pointer $\vtPtr\in\vtPtrSet=\aSet{0,\ldots|s|}$
that is initially positioned before the first letter of the input string $s$,
~$\vtPtr\!=\!0$,
and then increased with the reading of $s$
until it reaches the position after the last letter,
$\,\vtPtr\!=\!|s|$.
At any moment, $\vtPtr$ equals the length of the prefix of $s$
that has already been read.

As it is usual for the Viterbi algorithm,
we use a trellis $\vtNodeSet\!=Q\times\vtPtrSet$,
consisting of nodes $\vtNode\!=\!\aTuple{q,\vtPtr}$
which express that a state $q\!\in\!Q$ is reached after reading $\vtPtr$ letters of $s$
(Figure~\ref{fig:vit-1tape}).
We divide the trellis into several node sets
$\vtNodeSet_{\vtPtr} = \aSet{\vtNode\!=\!\aTuple{q,\vtPtr}} \subseteq \vtNodeSet$,
each corresponding to a pointer position $\vtPtr$ or to a column of the trellis.
For each node $\vtNode$,
we maintain three variables referring to $\vtNode$'s best prefix:
	$\vtPreWgt_{\vtNode}$ being its weight,
	$\vtPreNode_{\vtNode}$ its last node (immediately preceding $\vtNode$), and
	$\vtPreArc_{\vtNode}$ its last transition $e\!\in\!E$ of $A\tapnum{1}$.
The $\vtPreNode_{\vtNode}$ are back-pointers
that fully define the best prefix of each node $\vtNode$.
All $\vtPreWgt_{\vtNode}$, $\vtPreNode_{\vtNode}$, and $\vtPreArc_{\vtNode}$
are initially undefined ($\,=\!\Null\,$).\footnote{
  The variables $\vtPreWgt_{\vtNode}$, $\vtPreNode_{\vtNode}$, and $\vtPreArc_{\vtNode}$
  can be formally regarded as elements of the vectors
  $\vtPreWgtArr$, $\vtPreNodeArr$, and $\vtPreArcArr$,
  respectively, that are indexed by values of $\vtNode$.
  In a practical implementation is, however, meaningful to store these variables
  directly on the node that they refer to.
}


\begin{figure}[htb]
  \begin{center}
    {\small \input{vit-1tape.pc} }
    \vspace{-2ex}
    \caption{Pseudocode of $1$-tape best-path search
      \label{pc:vit-1tape}}
  \end{center}
\end{figure}

\subsection{Algorithm}

The algorithm \FUNCT{FsaViterbi}{~}
returns from all paths $\path$ of the $1$-WFSM $A\tapnum{1}$ that accept the string $s$,
the one with minimal weight
(Figure~\ref{pc:vit-1tape}).
$A\tapnum{1}$ must not contain any transitions labeled with $\eps$ (the empty string).
At least a partial order must be defined on the semiring of weights.
Nothing else is required concerning the labels, weights, or structure of $A\tapnum{1}$.\footnote{
  Cycles are, e.g., not required to have non-negative weights (as for Dijkstra's algorithm)
  because all paths of interest are constrained by the input string.
}

The algorithm starts with creating an initial node set $\vtNodeSet_{\sf initial}=\vtNodeSet_0$
for the initial position $\vtPtr=0$ of the reading pointer.
The set $\vtNodeSet_{\sf initial}$ contains a node for each initial state of $A\tapnum{1}$
(Lines~\ref{pc1:L101}--\ref{pc1:L104}).
The prefix weights $\vtPreWgt_\vtNode$ of these nodes are set to the initial weight $\wgtInit(q)$
of the respective states $q$.
The set of node sets $\vtNodeSetSet$ contains only $\vtNodeSet_{\sf initial}$ at this point
(Line~\ref{pc1:L105}).

In the subsequent iteration (Lines~\ref{pc1:L201}--\ref{pc1:L215}),
reaching from the first to the one but last pointer position, $p=0,\ldots|s|\!-\!1$,
we inspect all outgoing transitions $e\!\in\!E(q)$
of all states $q\!\in\!Q$
for which there is a node $\vtNode\!=\!\aTuple{q,\vtPtr}$ in $\vtNodeSet_{\vtPtr}$.
If the label $\lab(e)$ of $e$ matches $s$ at position $p$,
we create a new node $\vtNode^\prime=\aTuple{\eTrg(e),{\vtPtr^\prime}}$
for the target $\eTrg(e)$ of $e$ (Line~\ref{pc1:L204}).
Its prefix weight $\vtPreWgt^\prime$ equals the current node's weight $\vtPreWgt_{\vtNode}$
multiplied by the weight $w(e)$ of $e$.
The node set $\vtNodeSet_{\vtPtr^\prime}$ for the new $\vtNode^\prime$
is created and inserted into the set of node sets $\vtNodeSetSet$
~(if it does not exist yet; Line~\ref{pc1:L211}).
Then $\vtNode^\prime$ is inserted into $\vtNodeSet_{\vtPtr^\prime}$
~(if it is not yet a member of it; Line~\ref{pc1:L213}).
If the prefix weight of $\vtNode^\prime$ is still undefined,
$\vtPreWgt_{\vtNode^\prime}=\Null$
~(because no prefix of $\vtNode^\prime$ has been analyzed yet),
or if it is higher than the weight of the currently analyzed new prefix,
$\vtPreWgt_{\vtNode^\prime} > \vtPreWgt^\prime$,
then the variables $\vtPreWgt_{\vtNode^\prime}$, $\vtPreNode_{\vtNode^\prime}$,
and $\vtPreArc_{\vtNode^\prime}$ of $\vtNode^\prime$
are assigned values of the new prefix (Lines~\ref{pc1:L214}--\ref{pc1:L215}).

The algorithm terminates by selecting the node $\widehat\vtNode$,
corresponding to the path with the minimal weight,
from the final node set $\vtNodeSet_{\sf final}=\vtNodeSet_{|s|}$.
This weight is the product of the node's prefix weight $\vtPreWgt_{\vtNode}$
and the final weight $\wgtFin(q)$ of the corresponding state $q\!\in\!Q$
(Line~\ref{pc1:L301}).
The function \FCT{getPath}{~} identifies the best path $\path$
by following all back-pointers $\vtPreNode_{\vtNode}$,
from the node $\widehat\vtNode\in\vtNodeSet_{\sf final}$
to some node $\vtNode\in\vtNodeSet_{\sf initial}$,
and collecting all transitions $e\!=\!\vtPreArc_{\vtNode}$ it encounters.
Finally, $\path$ is returned.


\subsection{$\eps$-Transitions}

The algorithm can be extended to allow for $\eps$-transitions (but not for $\eps$-cycles).
The source and target node, $\vtNode$ and $\vtNode^\prime$, of an $\eps$-transition
would be in the same $\vtNodeSet_{\vtPtr}$.
If $\vtNode^\prime\!=\!\aTuple{q^\prime, p^\prime}$ is actually inserted into $\vtNodeSet_{\vtPtr}$
(Line~\ref{pc1:L213})
or if its variables $\vtPreWgt_{\vtNode^\prime}$, $\vtPreNode_{\vtNode^\prime}$,
and $\vtPreArc_{\vtNode^\prime}$ change their values
(Lines~\ref{pc1:L214}--\ref{pc1:L215}),
then we have to (re-)``include'' $\vtNode^\prime$ into the iteration over all nodes
of the currently inspected $\vtNodeSet_{\vtPtr}$
(Line~\ref{pc1:L204}).
The algorithm will still terminate
since there can be only finite sequences of $\eps$-transitions
(as long as we have no $\eps$-cycles).


\subsection{Best transduction}

The algorithm \FUNCT{FsaViterbi}{~}
can be used for compiling the best transduction of a given input string $s$
by a $2$-WFSM (weighted transducer).
For this, we identify the best path $\path$ accepting $s$ on its input tape
and take the label of $\path$'s output tape as best output string $v$.


\section{$n$-Tape Best-Path Search
  \label{sec:vit-ntape}}

We come now to the central topic of this paper:
the generalization of the Viterbi algorithm
for searching for the best of all paths of an $n$-WFSM, $A\tapnum{n}$,
that accept a given $n$-tuple of input strings,
$s\tapnum{n}\!=\!\aTuple{s_1,\ldots s_n}$.
This requires relatively few modifications to the above explained
structures and algorithm (Section~\ref{sec:vit-1tape}).


\subsection{Structures}

The main difference wrt. the previous structures is that now
our reading pointer is a vector of $n$ natural integers,
$
  \vtPtr\tapnum{n}\!=\!\aTuple{\vtPtr_1,\ldots\vtPtr_n}  \in
  \left(\vspc{2ex} \aRange{0,\dots|s_1|} \times\ldots
  \times \aRange{0,\dots|s_n|} \;\right)
  \subset \Nat^n
$.
The pointer is initially positioned before the first letter 
of each $s_i$ ~($\forall i\!\in\!\aRange{1,n}$),
~$\,\vtPtr\tapnum{n}\!=\!\aTuple{0,\ldots 0}\,$.
Its elements $p_i$ are then increased according to the non-synchronized reading of the $s_i$
on the tapes $i$ ~($\forall i\!\in\!\aRange{1,n}$),
until the pointer reaches its final position after the last letter of each $s_i$,
~$\,\vtPtr\tapnum{n}\!=\!\aTuple{|s_1|,\ldots |s_n|}\,$.

More precisely, a pointer is an element of the monoid $\aTuple{\Nat^n, +, {\bf 0}}$
with $+$ being vector addition and ${\bf 0}$ the vector of $n$ $0$'s.
We have a partial order of pointers.
Let $ \vtLess\; : \Nat^n\!\times\!\Nat^n \rightarrow \aSet{\srTrue, \srFalse} $.
Let $ a,b\in\Nat^n $,
then $ a \vtLess b  \biimplies  \left(\; \exists c\in\Nat^n, c\not={\bf 0} : a+c=b \right)\; $.
We say $a$ {\it precedes} $b$.
It holds that $ a \vtLess b  \implies  \left(\; \sum_{i=1}^n a_i < \sum_{i=1}^n b_i \right)\; $
where $a_i$ and $b_i$ are the vector elements.

\begin{figure}[ht]
  \begin{center}
    \includegraphics[scale=0.5,angle=0]{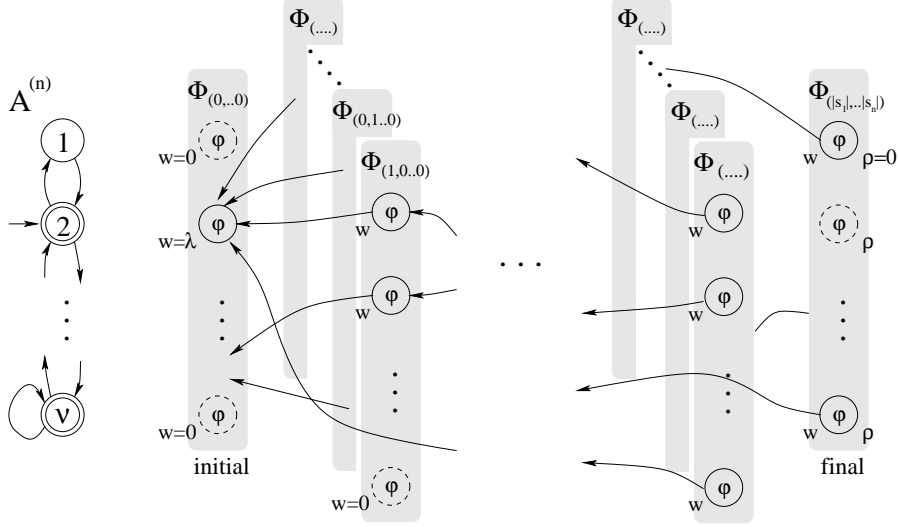}
    \caption{Modified trellis for $n$-tape best-path search
	\label{fig:vit-ntape}}
  \end{center}
\end{figure}

In the trellis (Figure~\ref{fig:vit-ntape})
we have still one node set $\vtNodeSet_{\vtPtr\tapnum{n}}$
per pointer position $\vtPtr\tapnum{n}$,
a single initial node set $\vtNodeSet_{\sf initial}\!=\!\vtNodeSet_{\aTuple{0,\dots 0}}$
and a single final node set $\vtNodeSet_{\sf final}\!=\!\vtNodeSet_{\aTuple{|s_1|,\dots|s_n|}}$.
There are, however, several nodes sets in parallel between the two
(corresponding to pointers $\vtPtr\tapnum{n},{\vtPtr^\prime}\tapnum{n}$
 not preceding each other, i.e.,
 $\vtPtr\tapnum{n}\!\not\vtLess\!{\vtPtr^\prime}\tapnum{n}  \logAnd
  {\vtPtr^\prime}\tapnum{n}\!\not\vtLess\!\vtPtr\tapnum{n}$).

\subsection{Algorithm}

The algorithm \FUNCT{FsmViterbi}{~}
returns from all paths $\path\tapnum{n}$ of the $n$-WFSM $A\tapnum{n}$
that accept the string tuple $s\tapnum{n}$, the one with minimal weight
(Figure~\ref{pc:vit-ntape}).
$A\tapnum{n}$ must not contain any transitions labeled with $\aTuple{\eps,\ldots\eps}$.\footnote{
  The algorithm can be extended to allow for $\aTuple{\eps,\ldots\eps}$-transitions
  (but not for $\aTuple{\eps,\ldots\eps}$-cycles)
  as described in Section~\ref{sec:vit-1tape}.
}

The initial node set $\vtNodeSet_{\sf initial}\!=\!\vtNodeSet_{\aTuple{0,\dots 0}}$
is created as before, and inserted into the set of node sets $\vtNodeSetSet$
(Lines~\ref{pc2:L101}--\ref{pc2:L105}).
In addition, it is inserted into a Fibonacci heap\footnote{
  Alternatively, one could use a binary heap.
  Tests on a concrete example have, however, shown that the algorithm performs slightly better
  with a Fibonacci heap
  (Table~\ref{tab:AlignResults}).
}
$\vtNodeSetHeap$
~(Line~\ref{pc2:L105})
\cite{fredman+tarjan:1987}.
This heap contains node sets $\vtNodeSet_{\vtPtr\tapnum{n}}$
that have not yet been processed,
and uses $\sum_{i=1}^n \vtPtr_i$ as sorting key.

The subsequent iteration continues as long as $\vtNodeSetHeap$ is not empty
(Lines~\ref{pc2:L201}--\ref{pc2:L215}).
The function \FCT{extractMinElement}{~}
extracts the (or a) minimal element $\vtNodeSet_{\vtPtr\tapnum{n}}$ from $\vtNodeSetHeap$
~(Line~\ref{pc2:L202}).
Due to our sorting key,
none of the remaining $\vtNodeSet_{{\vtPtr^\prime}\tapnum{n}}$ in $\vtNodeSetHeap$
is a predecessor to $\vtNodeSet_{\vtPtr\tapnum{n}}$~:~
$
\forall\vtNodeSet_{{\vtPtr^\prime}\tapnum{n}}\!\in\!\vtNodeSetHeap \,,\;
{\vtPtr^\prime}\tapnum{n}\!\not\vtLess\!\vtPtr\tapnum{n}
$.
This property prevents the compilation of suffixes
of a $\vtNodeSet_{\vtPtr\tapnum{n}}$ that has some not yet analyzed prefixes
(which could lead to wrong choices).
The extracted $\vtNodeSet_{\vtPtr\tapnum{n}}$ is
handled almost as in the previous algorithm (Figure~\ref{pc:vit-1tape}).
Transition labels $\lab(e\tapnum{n})$ are required to match with a factor of $s\tapnum{n}$
at position $\vtPtr\tapnum{n}$
(Line~\ref{pc2:L206}).
New $\vtNodeSet_{{\vtPtr^\prime}\tapnum{n}}$ are inserted both into $\vtNodeSetSet$
and $\vtNodeSetHeap$
~(Lines~\ref{pc2:L210}--\ref{pc2:L211}).

\begin{figure}[ht]
  \begin{center}
    {\small \input{vit-ntape.pc} }
    \vspace{-2ex}
    \caption{Pseudocode of $n$-tape best-path search
      \label{pc:vit-ntape}}
  \end{center}
\end{figure}


\subsection{Best transduction}

The algorithm \FUNCT{FsmViterbi}{~}
can be used for obtaining from a weighted $(n\!+\!m)$-WFSM (transducer)
with $n$ input and $m$ output tapes,
the best transduction of a given input $n$-tuple $s\tapnum{n}$.
For this, we identify the best path $\path\tapnum{n\!+\!m}$
accepting $s\tapnum{n}$ on its $n$ input tapes
and take the label of $\path$'s $m$ output tapes as best output $m$-tuple $v\tapnum{m}$.
Input and output tapes can be in any order.


\subsection{Complexity}

The trellis (Figure~\ref{fig:vit-ntape})
consists of at most $|\vtPtrSet|=\prod_{i=1}^{n}(|s_i|+1)$
node sets $\vtNodeSet_{\vtPtr\tapnum{n}}\!\in\!\vtNodeSetSet$.
Assuming approximately equal length $|s|$ for all $s_i$ of $s\tapnum{n}$,
we can simplify: $|\vtPtrSet|\approx(|s|+1)^n$.
For each node set $\vtNodeSet_{\vtPtr\tapnum{n}}$
we have to create at most $|Q|$ nodes $\vtNode\!\in\!\vtNodeSet_{\vtPtr\tapnum{n}}$,
which leads to a $\complexity\left( |s|^n |Q| \right)$ space complexity
for our algorithm.

Each $\vtNodeSet_{\vtPtr\tapnum{n}}$ is extracted once from the Fibonacci heap $\vtNodeSetHeap$
in $\complexity(\log|P|)$ time.
We analyze for $\vtNodeSet_{\vtPtr\tapnum{n}}$ at most $|E|$ transitions $e\!\in\!E$
of $A\tapnum{n}$.
For the target of each $e$ we find a $\vtNodeSet_{{\vtPtr^\prime}\tapnum{n}}\!\in\!\vtNodeSetSet$
in $\complexity(\log|P|)$ time
and a node $\vtNode^\prime\!\in\!\vtNodeSet_{{\vtPtr^\prime}\tapnum{n}}$
in $\complexity(\log|Q|)$ time.
Thus, \FUNCT{FsmViterbi}{~} has a worst-case overall time complexity of
$
\complexity\left(\; |P| (\log|P| + |E| (\log|P| + \log|Q|)) \;\right)
= \complexity\left(\, |P||E|\log|P||Q| \,\right)
= \complexity\left(\, |s|^n|E|\log|s|^n|Q| \,\right)
$~.

An HMM has exactly one transition per state pair, so that $|E|\!=\!|Q|^2$,
and an arity of $n\!=\!1$.
There would also be never more than one $\vtNodeSet_{\vtPtr\tapnum{n}}$ on the heap,
extractable in constant time.
In this case, our algorithm has a $\complexity\left( |s| |Q| \right)$ space
and a $\complexity\left( |s| |Q|^2 \right)$ time complexity,
as has the classical version of the Viterbi algorithm
(Section~\ref{sec:prelim}).


\section{Example: Word Alignment
  \label{sec:align}}

In this section we illustrate our $n$-tape best path search on a practical example:
the alignment of word pairs.

Suppose, we want to create a (non-weighted) transducer, $D\tapnum{2}$,
from a list of word pairs $s\tapnum{2}$
of the form $\aTuple{\WordD{inflected form}, \WordD{lemma}}$,
e.g., $\aTuple{\Word{swum}, \Word{swim}}$,
such that each path of the transducer is labeled with one of the pairs.
We want to use only transition labels of the form
$\aTuple{\sigma,\sigma}$, $\aTuple{\sigma,\eps}$, or $\aTuple{\eps,\sigma}$ ~($\forall\sigma\in\Sigma$),
while keeping paths as short as possible.
For example,
$\aTuple{\Word{swum}, \Word{swim}}$ should be encoded either by the sequence
$\aTuple{\Word{s},\Word{s}}\aTuple{\Word{w},\Word{w}}%
\aTuple{\Word{u},\eps}\aTuple{\eps,\Word{i}}\aTuple{\Word{m},\Word{m}}$
or by
$\aTuple{\Word{s},\Word{s}}\aTuple{\Word{w},\Word{w}}%
\aTuple{\eps,\Word{i}}\aTuple{\Word{u},\eps}\aTuple{\Word{m},\Word{m}}$,
rather than by the ill-formed
$\aTuple{\Word{s},\Word{s}}\aTuple{\Word{w},\Word{w}}%
\aTuple{\Word{u},\Word{i}}\aTuple{\Word{m},\Word{m}}$,
or the sub-optimal  
$\aTuple{\Word{s},\eps}\aTuple{\Word{w},\eps}\aTuple{\Word{u},\eps}\aTuple{\Word{m},\eps}%
\aTuple{\eps,\Word{s}}\aTuple{\eps,\Word{w}}\aTuple{\eps,\Word{i}}\aTuple{\eps,\Word{m}}$.
To achieve this, we perform for each word pair an alignment based on minimal edit distance.


\subsection{Standard solution with edit distance matrix}

A well known standard solution for word alignment is based on edit distance
which is a string similarity measure
defined as the minimum cost needed to convert one string into another
\cite{wagner+fischer:1974,pirkola+al:2003}.

\begin{figure}[ht]
  \vspace{1ex}
  \begin{center}
    \begin{minipage}{65mm}
      \begin{center}
	\includegraphics[scale=0.5,angle=0]{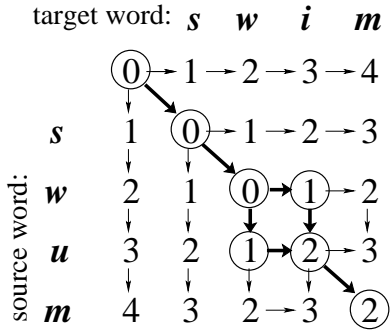}

	\caption{Edit distance matrix
		$X\!=\!\{x_{i,j}\}$
		(choices are indicated by arrows; minimum cost paths by thick arrows and circles)
		\label{fig:EditDistanceMatrix}}
      \end{center}
    \end{minipage}
    \spc{10mm}
    \begin{minipage}{60mm}
      \begin{center}
	{\small
	  \input{align.pc}
	}

	\vspace{-3ex}
	\caption{Pseudocode of compiling an {\it edit distance matrix}
		\label{pc:EditDistanceMatrix}}
      \end{center}
    \end{minipage}
  \end{center}
\end{figure}

For two words, $a\!=\!a_1\ldots a_n$ and $b\!=\!b_1\ldots b_m$,
the edit distance can be compiled with a matrix
$X\!=\!\{x_{i,j}\}$ ~($i\!\in\!\aRange{0,n}$, $j\!\in\!\aRange{0,m}$)
(Figures~\ref{fig:EditDistanceMatrix} and~\ref{pc:EditDistanceMatrix}).
A horizontal move in $X$ at a cost $c_I$ expresses an {\it insertion}\/,
a vertical move at a cost $c_D$ a {\it deletion}\/,
and a diagonal move at a cost $c_S$ a {\it substitution}\/ if $a_i\!\not=\!b_j$
or no edit operation if $a_i\!=\!b_j$.
We set $c_I\!=\!c_D\!=\!1$,
$c_S\!=\!\infty$ for $a_i\!\not=\!b_j$ (to disable substitutions),
and $c_S\!=\!0$ for $a_i\!=\!b_j$.
The element $x_{0,0}$ is set to $0$ and all other $x_{i,j}$ to
$\min(x_{i,j-1}+c_I\,,\; x_{i-1,j}+c_D\,,\; x_{i-1,j-1}+c_S)$,
insofar as these choices are available,
proceeding top-down and left-to-right.
The choices made to go from $x_{0,0}$ to $x_{n,m}$ describe the set of paths with (the same) minimal cost.
Each of these paths defines a sequence of edit operations for transforming $a$ into $b$.

The algorithm operates in $\complexity(|a||b|)$ time and space complexity.

\subsection{Solution with 2-tape best path search}

Alternatively, word alignment can be performed by best path search on an $n$-WFSM,
such as $A\tapnum{5}$ generated from the expression
\cite{isabelle+kempe:2004}
%
\begin{eqnarray}
  A\tapnum{5} & \;=\; & \left(\;
			\aTuple{\aTuple{\Any,\Any,\Any,\Any,\algnK}_{\aSet{1=2=3=4}} , 0}
			 \right.	 \nonumber \\
		& &	 \left. \spc{5ex}
			\;\cup\; \aTuple{\aTuple{\eps,\Any,\algnEps,\Any,\algnI}_{\aSet{2=4}} , 1}
			 \;\cup\; \aTuple{\aTuple{\Any,\eps,\Any,\algnEps,\algnD}_{\aSet{1=3}} , 1} \;\right)^*
	\label{eq:align}
\end{eqnarray}

\noindent
where $\Any$~ can be instantiated by any symbol $\sigma\!\in\!\Sigma$,
$\algnEps$ is a special symbol representing $\eps$ in an alignment,
$\aSet{1\!=\!2\!=\!3\!=\!4}$ a constraint
requiring the $\Any$'s on tapes $1$ to $4$ to be instantiated by the same symbol
\cite{nicart+al:2006a},\footnote{
  Roughly following \cite{kempe+champarnaud+eisner:2004a},
  we employ here a simpler notation for constraints than in~\cite{nicart+al:2006a}.
}
and $0$ and $1$ are weights over the semiring $\aTuple{\srSetN\cup\aSet{\infty}, \min, +, \infty, 0}$.

Input word pairs $s\tapnum{2}\!=\!\aTuple{s_1,s_2}$ will be matched on tape 1 and 2,
and aligned output word pairs generated from tape 3 and 4.
A symbol pair $\aTuple{\Any,\Any}$ read on tape 1 and 2
is identically mapped to $\aTuple{\Any,\Any}$ on tape 3 and 4,
a $\aTuple{\eps,\Any}$ is mapped to $\aTuple{\algnEps,\Any}$,
and a $\aTuple{\Any,\eps}$ to $\aTuple{\Any,\algnEps}$.
$A\tapnum{5}$ will introduce $\algnEps$'s in $s_1$ (resp. in $s_2$) at positions
where $D\tapnum{2}$ shall have $\aTuple{\eps,\sigma}$-
(resp. a $\aTuple{\sigma,\eps}$-) transitions.
(Later, we simply replace in $D\tapnum{2}$ all $\algnEps$ by $\eps$.)

Thus, we obtain the full set of all possible alignments between $s_1$ and $s_2$.
The best alignment is the one with the lowest weight.
For example, $\aTuple{\Word{swum}, \Word{swim}}$ is mapped to a set of alignments,
including the two best ones,
$\aTuple{\Word{sw\algnEps um}, \Word{swi\algnEps m}}$
and $\aTuple{\Word{swu\algnEps m}, \Word{sw\algnEps im}}$, with weight 2 both.
The (or a) best alignment can be found without generating all alignments,
by means of our $n$-tape best path search (with $n\!=\!2$).

So far, we did not use tape 5.
It can serve for excluding certain paths.
For example, joining $A\tapnum{5}$ on tape 5 with $C\tapnum{1}$
\cite{kempe+al:2005a,kempe+al:2005b}
built from the expression $\neg(\Any^*\;\algnI\;\algnD\;\Any^*)$,
prohibiting an insertion ($\algnI$) to be immediately followed by a deletion ($\algnD$),
would leave only $\aTuple{\Word{swu\algnEps m}, \Word{sw\algnEps im}}$ as a best path.


The 5-WFSM from Equation~\eqref{eq:align}
has 1 state and 3 transitions.
Input is read on 2 tapes.
Our algorithm works on this example 
with a worst-case time complexity of
$
  \complexity(\, |s_1||s_2|\cdot 3\cdot \log(|s_1||s_2|\cdot 1) \,)
  = \complexity(\,|s_1||s_2|\log|s_1||s_2|\,)
$
and a worst-case space complexity of
$
  \complexity(\,|s_1||s_2|\cdot 1 \,)
  = \complexity(\,|s_1||s_2|\,)
$~.


\subsection{Test results}

We tested our $n$-tape best-path algorithm on the alignment of the German word pair
$\aTuple{\Word{gemacht}, \Word{machen}}$ ~(English: $\aTuple{\WordD{done}, \WordD{do}}$),
leading to  
$\aTuple{\Word{gemacht\algnEps\algnEps}, \Word{\algnEps\algnEps{mach}\algnEps{en}}}$.
We repeated this test for the word pairs $\aTuple{s_1^r, s_2^r}$
with $s_1=$``\Word{gemacht}'' and $s_2$=``\Word{machen}'',
and $r\!\in\!\aRange{1,8}$.\footnote{
  For example, for $r\!=\!2$ we have
  $\aTuple{{\sf\small gemachtgemacht}, {\sf\small machenmachen}}$.
}

\def\S{\spc{1.1ex}}
\def\D{\spc{0.6ex}}

\begin{table}[ht]
  \vspace{1ex}
  \begin{center}
    \begin{math} 
      \begin{tabular}{ c | r r r | c } \hline
	\spc{2ex}$r$\spc{2ex}	& \spc{2ex}A\spc{0.5ex}	& \spc{3ex}B\spc{2.5ex}	& \spc{3ex}C\spc{2.8ex}	& \spc{4ex}D\spc{3ex}	\\ \hline
	1	&  1	&   1\D\S\S	&   1\D\S\S	& 1.056	\\
	2	&  4	&   4.12	&   5.48	& 1.041	\\
	3	&  9	&   9.41	&  14.3\S	& 1.057	\\
	4	& 16	&  17.1\S	&  27.9\S	& 1.029	\\
	5	& 25	&  27.2\S	&  46.5\S	& 1.059	\\
	6	& 36	&  39.8\S	&  70.5\S	& 1.016	\\
	7	& 49	&  54.1\S	& 100\D\S\S	& 1.005	\\
	8	& 64	&  70.8\S	& 135\D\S\S	& 1.006	\\ \hline
      \end{tabular}
    \end{math}

    \caption{Test results for word pair alignment with 2-tape best path search
	\label{tab:AlignResults}}
  \vspace{-1ex}
  \end{center}
\end{table}

\noindent
The columns of Table~\ref{tab:AlignResults} show for different $r$~:
\begin{itemize}
\item[(A)]
  an estimated time ratio of $r^2$ for the classical approach with an edit distance matrix,
\item[(B)]
   the measured time ratio for 2-tape best path search (wrt. 3.93 milliseconds for $r=1$)
   using a Fibonacci heap,
\item[(C)]
  an estimated worst-case time ratio of
  $
  \frac{(7r\cdot 6r) \log (7r\cdot 6r)}{(7\cdot 6) \log (7\cdot 6)}
  = r^2(1\!+\!2\frac{\log r}{\log 42})
  $
  corresponding to the worst-case complexity of $\complexity(7r 6r \log 7r 6r)$
  for the two words of length $7r$ and $6r$, respectively, and
\item[(D)]
  the measured time increase factor when using a binary instead of a Fibonacci heap.
\end{itemize}

Comparing the columns A and B shows a time complexity slightly above
$\complexity(r^2) = \complexity(\,|s_1^r||s_2^r|\,)$,
being much lower than the worst-case time complexity in column C,
for our algorithm on this example.

\pagebreak


\section{An Alternative Approach
  \label{sec:alternatives}}

A well-known straight forward alternative to the above $n$-tape best-path search
on an $n$-WFSM $A\tapnum{n}$ is
to intersect $A\tapnum{n}$ with an $n$-WFSM $I\tapnum{n}$,
containing a single path labeled with the input $n$-tuple $s\tapnum{n}$,
and then to apply a classical shortest-distance algorithm, ignoring the labels.


\subsection{Intersection}

The intersection $B\tapnum{n} = I\tapnum{n} \cap A\tapnum{n}$
can be compiled as the join $I\tapnum{n} \JOIN{1=1,\ldots n=n} A\tapnum{n}$
\cite{kempe+champarnaud+eisner:2004a}.
In general, it has undecidable emptiness and rationality \cite{rabin+scott:1959}.
In our case, however,
with $A\tapnum{n}$ being $\aTuple{\eps,\ldots\eps}$-cycle free
and $I\tapnum{n}$ acyclic,
it is even for non-commutative semirings always rational.\footnote{
  The intersection of two $n$-WFSM over non-commutative semirings
  is in general not rational (even for $n\!=\!1$).
}

Actually, the trellis $\vtNodeSet$ in Figure~\ref{fig:vit-ntape}
corresponds partially to $B\tapnum{n}$.
Each node $\vtNode\!\in\!\vtNodeSet$
corresponds to a state $q\!\in\!Q_B$ of $B\tapnum{n}$
(and vice versa);
however, only those transitions $e\!\in\!E_B$ of $B\tapnum{n}$
that correspond to a state's best prefix,
occur as ``best transitions'' $e_{\vtNode}$ in $\vtNodeSet$.\footnote{
  Due to this analogy, one can easily derive an $n$-tape intersection (or join) algorithm,
  for precisely our case, from the algorithm in Figure~\ref{pc:vit-ntape}.
  Trellis nodes would become states of the resulting $n$-WFSM.
  All of their incoming transitions would be constructed,
  rather than only those that correspond to a best prefix.  
  The state set would be partitioned like the trellis.
  The Fibonacci heap can be replaced by a stack
  (which does not decrease the overall time complexity),
  because the order in which partitions are treated would be irrelevant.
}

From this analogy we deduce that compiling the intersection $B\tapnum{n}$
has a worst-case time and space complexity of
$\complexity\left(\, |P||E|\log|P||Q| \,\right)$, with $|P|\!=\!(|s|+1)^n$,
equal to the time complexity for constructing the trellis.
The result, $B\tapnum{n}$, has at most
$\nu \leq |P||Q|$ states and $\mu \leq |P||E|$ transitions.


\subsection{Shortest-distance algorithms}

Since any $n$-WFSM with multiple initial states can be transformed
into one with a single initial state,
we can use any algorithm that solves a single-source shortest-distance problem,
such as Dijkstra's algorithm \cite{dijsktra:1959}
combined with Fibonacci heaps \cite{fredman+tarjan:1987},
that operates in $\complexity(\mu + \nu\log\nu)$ time,
or Bellman-Ford's algorithm \cite{bellman:1958,ford+fulkerson:1956}
operating in $\complexity(\mu\nu)$ time,
with $\nu$ being the number of states and $\mu$ the number of transitions.

Recently, it has been shown that any single-source shortest-distance algorithm on 
directed graphs has a lower bound of $\Omega(\mu + \min(\nu\log\nu ,\; \nu\log\rho))$
where $\rho$ is the ratio of the maximal to minimal transition weight
\cite{pettie:2003}.
Since we cannot make any assumption concerning $\rho$ in general,
we consider $\widehat\Omega(\mu + \nu\log\nu)$ as a ``worst-case lower bound''.
It equals the upper bound of Dijkstra's algorithm.

On the intersection $B\tapnum{n} = I\tapnum{n} \cap A\tapnum{n}$,
Dijkstra's algorithm requires $\complexity(|P||E|+|P||Q|\log|P||Q|)$ time,
and Bellman-Ford's $\complexity(|P|^2|E||Q|)$ time, in the worst case.
The sets $E$ and $Q$ refer to $A\tapnum{n}$.


\subsection{Complete estimate}

Intersection and Dijkstra's algorithm have together
a worst-case time complexity of  \linebreak
$
\complexity\left(\, |P||E|\log|P||Q| + |P||E| + |P||Q|\log|P||Q| \,\right)
\approx \complexity\left(\, |P|(|E|+|Q|) \log|P||Q| \,\right)
$.
For intersection and Bellman-Ford's algorithm it is
$
\complexity\left(\, |P||E|\log|P||Q| + |P|^2|E||Q| \,\right) =
\complexity\left(\, |P||E|\,(|P||Q|\!+\!\log|P||Q|) \,\right)
$.
Both combinations exceed the complexity of our algorithm.

This result is not surprising since
only building the trellis $\vtNodeSet$ should take less time
than building the intersection $B\tapnum{n}$
(which is a kind of ``superset'' of $\vtNodeSet$)
and then performing a best-path search.


\pagebreak


\section{Conclusion
   \label{sec:conclusion}}

We presented an algorithm for identifying the path with minimal (resp. maximal) weight
in a given {\it $n$-tape weighted finite-state machine}\/ ($n$-WFSM), $A\tapnum{n}$,
that accepts a given $n$-tuple of input strings,
$s\tapnum{n}\!=\!\aTuple{s_1,\ldots s_n}$.
This problem is of particular interest because it allows us also
to compile the best transduction of a given input $n$-tuple $s\tapnum{n}$
by a weighted $(n\!+\!m)$-WFSM (transducer), $A\tapnum{n+m}$, with $n$ input and $m$ output tapes.
For this, we identify the best path accepting $s\tapnum{n}$ on its $n$ input tapes,
and take the label of its output tapes as best output $m$-tuple $v\tapnum{m}$.
(Input and output tapes can be in any order.)

Our algorithm is a generalization of the Viterbi algorithm
which is generally used for detecting the most likely path
in a {\it Hidden Markov Model}\/ (HMM)
for an observed sequence of symbols emitted by the HMM.
In the worst case,
it operates in $\complexity\left(\, |s|^n|E|\log|s|^n|Q| \,\right)$ time,
where $n$ and $|s|$ are the number and average length of the strings in $s\tapnum{n}$,
and $|Q|$ and $|E|$ the number of states and transitions of $A\tapnum{n}$,
respectively.

We illustrated our $n$-tape best path search on a practical example,
the alignment of word pairs (i.e., $n\!=\!2$),
and provided test results that show a time complexity slightly higher than
$\complexity\left(\, |s|^2 \,\right)$.

Finally, we discussed a straight forward alternative approach for solving our problem,
that consists in intersecting $A\tapnum{n}$ with an $n$-WFSM $I\tapnum{n}$,
that has a single path labeled with the input $n$-tuple $s\tapnum{n}$,
and then applying a classical shortest-distance algorithm, ignoring the labels.
This has, however, a worst-case time complexity of
$ \complexity\left(\, |s|^n(|E|+|Q|)\log|s|^n|Q| \,\right) $,
which is higher than that of our algorithm.



\bibliographystyle{fullname}

\bibliography{ntvit}


\end{document}